\documentclass[journal]{IEEEtran}

\usepackage{url}  
\usepackage{graphicx}  
\usepackage{amsmath,amssymb}
\usepackage{color}
\usepackage{epstopdf}
\usepackage{array}
\usepackage{multirow}
\usepackage{bbm}
\usepackage{amsfonts}
\usepackage{booktabs}
\usepackage{bm}
\usepackage{float}
\usepackage{subfigure}
\usepackage{cite}
\usepackage{ifpdf}
\usepackage{booktabs}

\usepackage{bbding}

\usepackage{algorithmic}
\usepackage{algorithm}

\usepackage[breaklinks=true,colorlinks,bookmarks=true]{hyperref}

\begin{document}

\title{Improving Network Interpretability via Explanation Consistency Evaluation}

\author{Hefeng Wu, Hao Jiang, Keze Wang, Ziyi Tang, Xianghuan He, Liang Lin,~\IEEEmembership{Fellow,~IEEE}
\thanks{This work was supported in part by National Natural Science Foundation of China (NSFC) under Grant No. 62272494, 62276283 and 62325605, in part by Guangdong Basic and Applied Basic Research Foundation under Grant No. 2023A1515012985, 2023A1515012845 and 2023A1515011374, and in part by Basic and Applied Basic Research Special Projects under Grant No. SL2022A04J01685. (Corresponding author: Keze Wang)}
\thanks{Hefeng Wu, Hao Jiang, Keze Wang, Ziyi Tang, Xianghuan He, and Liang Lin are with the School of Computer Science and Engineering, Sun Yat-sen University, Guangzhou, China (e-mail: wuhefeng@gmail.com, jiangh227@ mail2.sysu.edu.cn, kezewang@gmail.com, tangzy27@mail2.sysu.edu.cn, hexh 28@mail2.sysu.edu.cn, linliang@ieee.org).}
}

\markboth{IEEE Transactions on Multimedia}%
{Wu \MakeLowercase{\textit{et al.}}: }

\maketitle

\begin{abstract}

While deep neural networks have achieved remarkable performance, they tend to lack transparency in prediction. The pursuit of greater interpretability in neural networks often results in a degradation of their original performance. Some works strive to improve both interpretability and performance, but they primarily depend on meticulously imposed conditions. In this paper, we propose a simple yet effective framework that acquires more explainable activation heatmaps and simultaneously increase the model performance, without the need for any extra supervision. Specifically, our concise framework introduces a new metric, i.e., explanation consistency, to reweight the training samples adaptively in model learning. The explanation consistency metric is utilized to measure the similarity between the model's visual explanations of the original samples and those of semantic-preserved adversarial samples, whose background regions are perturbed by using image adversarial attack techniques. Our framework then promotes the model learning by paying closer attention to those training samples with a high difference in explanations (i.e., low explanation consistency), for which the current model cannot provide robust interpretations. Comprehensive experimental results on various benchmarks demonstrate the superiority of our framework in multiple aspects, including higher recognition accuracy, greater data debiasing capability, stronger network robustness, and more precise localization ability on both regular networks and interpretable networks. We also provide extensive ablation studies and qualitative analyses to unveil the detailed contribution of each component.

\end{abstract}


\IEEEpeerreviewmaketitle

\section{Introduction} \label{sec:intro}

\IEEEPARstart{T}{he} lack of transparency of deep convolutional neural networks (CNNs) has raised increasing concerns. Understanding these models' predictions has become an essential topic of study \cite{WangV23pami,LiLY24tmm,CuiWW19tmm,Wang00020tmm,PengYLH23TMM,Wang0WGZJW023iccv,vucnn}, where researchers make progress mainly in the following two directions. A straightforward way is network visualization, i.e., visualizing the learned filters/neurons inside networks after the training phase. Representative methods include gradient-based network visualization \cite{vucnn,vuit,gradcam,tdna} and inverted feature representation \cite{invertcnn}. However, since these methods only introduce constraints to generate visualizations with clear semantic meanings rather than chaotic results, the network itself is not further refined to provide better semantic explanations. Moreover, the reliability of these methods has been heavily questioned \cite{Ghorbani_2019,adebayo2018sanity}. In contrast to directly visualizing the well-trained networks, several methods \cite{icnn,ircnn,infogan,vae} attempt to allow the intrinsic interpretability of networks by encouraging their intermediate layers to learn interpretable features. These methods have a joint learning objective during the training to simultaneously increase both interpretability and discriminative power. However, as reported in \cite{qicnn}, the interpretability of the learned CNN features is not necessarily bound with its discriminative power. Therefore, these methods have to empirically balance a trade-off between achieving optimal accuracy and aligning the learned features with human-interpretable concepts.
Recent works \cite{zunino2020explainable,goyal2019counterfactual} attempt to address this dilemma. However, they either require extra fine-grained supervision \cite{zunino2020explainable} or get constrained by the sequential exhaustive search \cite{goyal2019counterfactual}.

\begin{figure*}[!t]
\centering
\includegraphics[width = 0.68\textwidth]{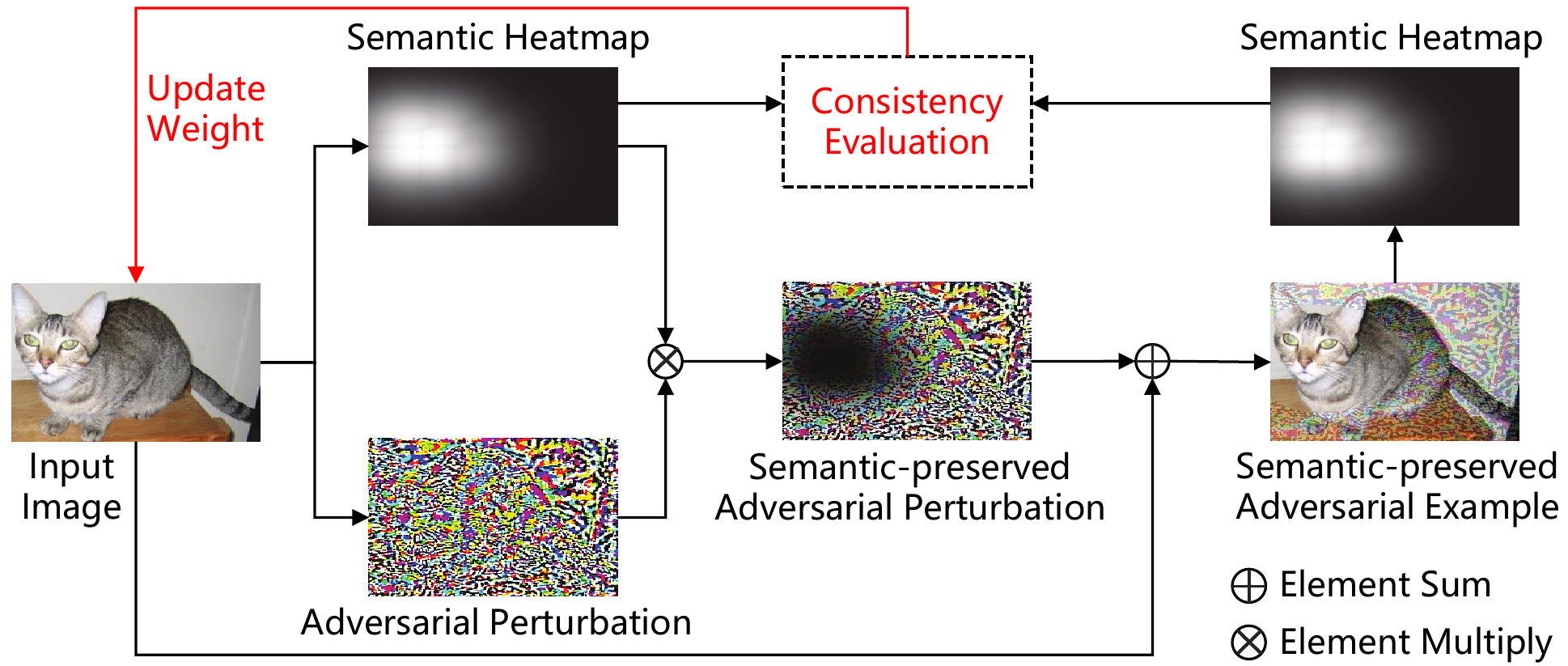}
\caption{Illustration of how the proposed semantic-preserved adversarial example is generated and how the explanation consistency is evaluated to update the sample weight of an input image in training.}
\label{fig:fusion}
\end{figure*}

In this paper, we seek for a simple and light-weight approach to address this problem. Specifically, as demonstrated in \cite{zhang2018examining}, biased training samples may encourage the networks to learn unreliable feature representations. This implies that the network interpretability is sensitive to the training samples in terms of their diversity and learning difficulty. Therefore, we hypothesize that refining the model training process of the existing methods may reduce the fragility of their generated explanations.
Inspired by the sample reweighting mechanism (e.g., hard negative mining \cite{ohem}, where more attention is paid to those hard samples) and self-paced learning \cite{spl,lin2017focal,ceal,SunXL023tmm,aspl} that progressively includes the samples from easy to hard into the training process, we argue that incorporating interpretability perspectives into the sample reweighting scheme is beneficial for addressing the shortcomings and undesirable trade-offs of existing methods. 

To this end, we introduce the concept of explanation consistency as a novel measure of explanation difficulty for samples and develop a simple yet effective evaluation strategy to quantify it. We further propose a concise framework to improve network interpretability by downgrading the training weights of the samples with high explanation consistency and increasing the weights of those with low explanation consistency during the training phase.

As illustrated in Figure~\ref{fig:fusion}, we first leverage the image adversarial attack techniques \cite{ehae} to generate the adversarial perturbation/noise from the given image. Instead of directly adding this noise into the whole image to mislead the network, we only perturb image regions that are less important according to the semantic region with high confidence produced by the network explanation technique. In this way, the perturbed adversarial example is regarded as semantic-preserved. Then, we feed this adversarial example into the network to make a second prediction and query for an explanation again. Intuitively, both these two predictions and their corresponding explanations should be consistent if the model accurately interprets this image and its semantic-preserved example for label prediction from similar semantic regions.
Otherwise, a higher weight should be assigned to this input image for further training. Note that, any adversarial examples are not included for network training. The reason is that although adversarial samples are semantic-preserving for humans, they are obviously not for the model. In contrast, our semantic-preserved adversarial samples should still be clear for the model if the explanations are faithful.
Although adversarial training is a fascinating topic for improving model robustness, the main focus of the paper is to improve the model interpretability by employing the adversarial techniques to generate semantic-preserved adversarial samples for measuring explanation consistency. Hence, we only reweight the training samples which can be correctly predicted by the model.

Through this unsupervised evaluation consistency checking, our framework can effectively improve the network interpretability by paying close attention to the training samples that are difficult for the network. These samples may reveal minor yet crucial differences for encouraging the filters/hidden units of the network to concentrate on extracting features from semantic regions of the input data. This is beneficial for both learning interpretable features and training discriminative classifiers. In such a hard-sample-high-priority progress, the network can gradually improve its interpretability as well as its discriminative power.

The main contributions of this paper are summarized as follows. First, we propose a concise yet effective framework to improve both network interpretability and performance. To the best of our knowledge, this paper is the first one to incorporate the explanation-based sample reweighting scheme into the training of networks to benefit the interpretability and discriminative power. Second, we develop a novel measurement for network explanation consistency by exploiting image adversarial attack techniques. Moreover, extensive experiments with comprehensive ablation analyses are conducted to verify the effectiveness of our framework.

\begin{figure*}[!t]
\centering
\includegraphics[width = 0.75\textwidth]{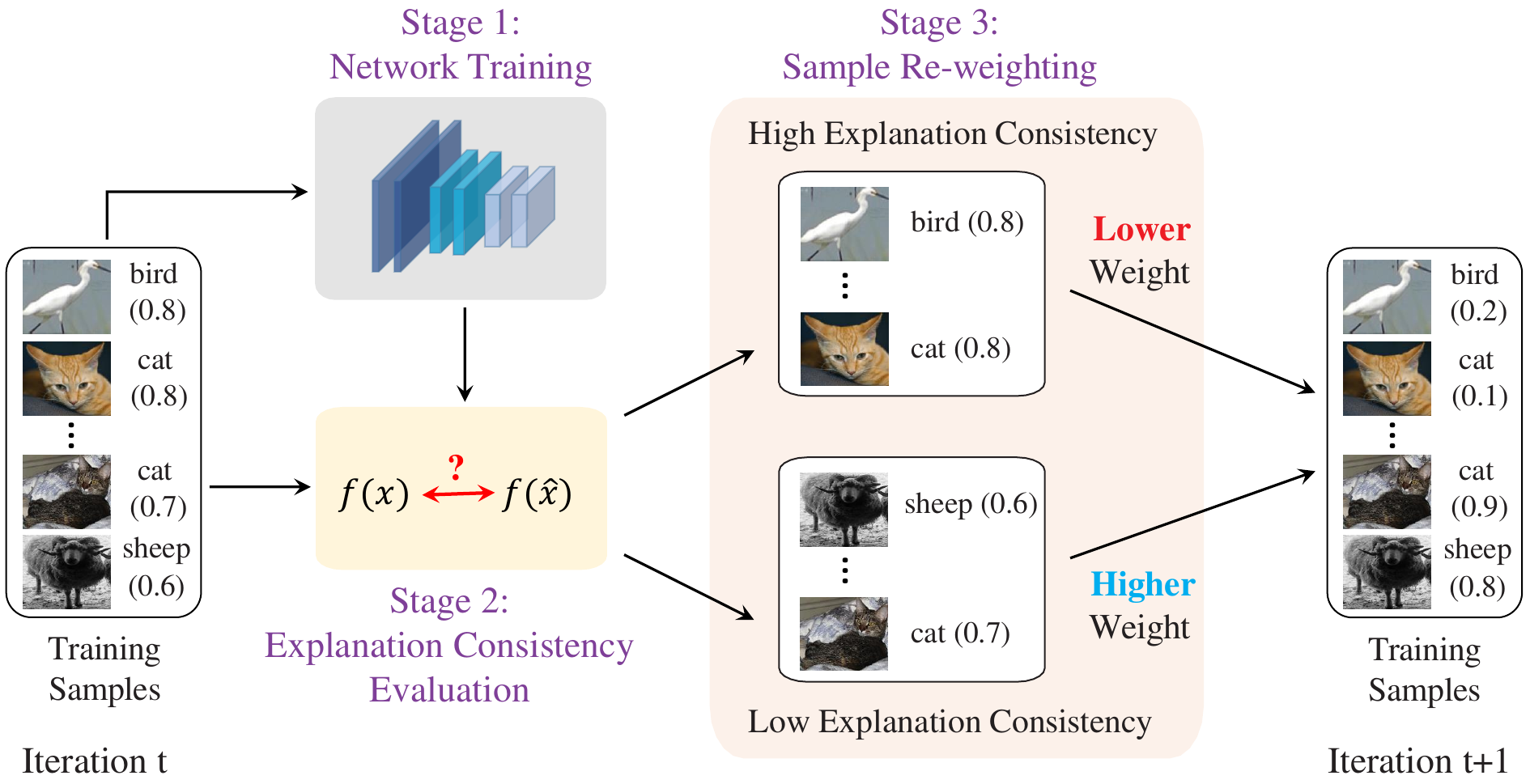}
\caption{The iterative training pipeline of the proposed framework that improves network interpretability by evaluating the explanation consistency. The arrows represent the work-flow. The category label and the training weight (numerical value) for each image are also illustrated.} 
\label{fig:framework}
\end{figure*}

The remainder of this paper is organized as follows. We review the related work in Section \ref{sec:related_work} and detailedly describe our framework in Section \ref{sec:Methodology}. The experimental evaluation and analysis are presented in Section \ref{sec:Experiments}, and Section \ref{sec:Conclusion} concludes the paper.

\section{Related Work}\label{sec:related_work}
\paragraph{Network Interpretability}
Generally, it is quite challenging to characterize the network interpretability. Efforts in addressing this objective can be roughly categorized into the following two lines of works. 
Mahendran and Vedaldi \cite{vuit} present an optimisation method to invert shallow and deep representations based on optimizing an objective function with gradient descent. 
Zeiler et al. \cite{vucnn} propose to employ deconvolutional networks to gives insight into the function of intermediate feature layers and the operation of the classifier. Dosovitskiy and Brox \cite{invertcnn} develop the up-convolutional nets to invert feature maps of convolutional layers into images. More recently, in a similar spirit, Zhang et al. \cite{tdna} present a way to model the top-down attention of a CNN classifier for generating task-specific attention maps.
Inspired by saliency methods \cite{vucnn,ZhuCW024aaai} that are widely used to highlight important input features, Ismail et al. \cite{IsmailBF21nips} introduce a saliency guided training procedure to improve the interpretability of deep neural networks.

\paragraph{Network Visualization} Aiming at explaining what image content/patches are emphasized by the network, a variety of methods \cite{WangV23pami,gradcam,integrad,idt,ennsq} focus on visualizing the layers/filters of the well-trained networks. For instance, Selvaraju et al. \cite{gradcam} introduce Grad-CAM, a gradient-based localization method for visualization. Grad-CAM generates a class activation map (CAM) highlighting the semantic image regions crucial for the prediction, without requiring changes to the network architecture or retraining. Sundararajan et al. \cite{integrad} propose to integrate all Vanilla gradients using a linear interpolation between a baseline input and the original input for visualization. 
Zhang et al. \cite{idt} propose to learn a decision tree to quantitatively explain the rationales of each prediction that is made by a pre-trained CNN. Chen et al. \cite{ennsq} present a method to explain the knowledge encoded in a CNN quantitatively and semantically by distilling knowledge. 
On the other hand, several methods concentrate on the extraction of pixel-level correlations between network inputs and outputs. Specifically, Lundberg and Lee \cite{imp} present a unified framework for interpreting predictions. The presented framework assigns each feature an important value for a particular prediction. Ribeiro et al. \cite{wsty} propose an explanation technique that explains the predictions of any classifier in an interpretable and faithful manner, by learning an interpretable model locally around the prediction. Rather than neglecting the training samples, Koh and Liang \cite{ubp} exploit the influence function, a classic technique in statistics, to understand the network predictions. However, as pointed out in \cite{benchmark}, the performances of all the aforementioned methods are not satisfactory enough, as the quality of their generated input feature importance may be worse or on par with a random assignment of importance.

\paragraph{Interpretable Networks} Instead of developing additional components or modules to interpret the learned feature representations of networks, enforcing the network to learn interpretable features is a new trend and has received increasing attention. Focusing on learning well-disentangled codes for generative networks, Chen et al. \cite{infogan} propose a generative adversarial network that also maximizes the mutual information between a small subset of the latent variables and the observation. To achieve a similar goal, Higgins et al. \cite{vae} reformulate the standard VAE framework as a constrained optimization problem with strong latent capacity constraint and independence prior pressures. Vaughan et al. \cite{ennaim} design a structured neural network especially to learn interpretable features via linear projections and univariate functions. Kindermans et al. \cite{hte} present a generalization for linear models to produce improved explanations for deep networks. Zhang et al. \cite{icnn} propose to clarify knowledge representations in high convolutional layers of the CNN. Wu and Song \cite{ircnn} present a method of formulating model interpretability in visual understanding tasks based on the idea of unfolding latent structures.

\paragraph{Explanation Evaluations} In terms of the masking or occlusion based evaluations of interpretability, both \cite{benchmark} and \cite{occlusion} are highly related to our proposed method. Specifically, Hooker et al. \cite{benchmark} propose to replace the fraction of the pixels estimated to be the most important with a fixed uninformative value. In this way, the existing interpretability methods can be evaluated by verifying how the accuracy of a retrained model degrades as the important pixels being removed. Samek et al. \cite{occlusion} focus on measuring how much small changes in the pixel value locally affect the network output. In contrast to the above-mentioned methods, our introduced strategy for evaluating explanation consistency concentrates on adversarially changing the value of the less important pixels to challenge the classifier.

\paragraph{Sample Reweighting}
Aimed at addressing the imbalance problem, sample reweighting is a common and popular strategy and has plenty of applications \cite{ohem,lin2017focal}. An example of this is hard negative mining (HNM), a technique that encourages the current model to concentrate on challenging samples by assigning them with higher weights. In \cite{ohem}, HNM has been shown to significantly improve the performance of deep learning models on object detection. In a similar spirit, Lin et al. \cite{lin2017focal} reshape the cross-entropy loss function such that it can down-weight to well-classified samples. Moreover, sample reweighting has also been employed by the recent robust learning techniques, i.e., self-paced learning \cite{spl,aspl,ceal}, to control the training progress by gradually including samples from easy to hard.

\section{Methodology}\label{sec:Methodology}

\subsection{Overview}
We introduce a simple yet effective framework to train the model, which not only produces more explainable activation heatmaps but also enhances prediction performance without additional supervision. 
Specifically, our framework is based on explanation consistency evaluation. Intuitively, it aligns with human recognition that when an image sample experiences some perturbation, but the semantic regions responsible for the network decision are essentially preserved, the explainable activation heatmaps should remain consistent, which we refer to as explanation consistency in this paper.
We utilize image adversarial attack techniques \cite{ehae} to generate perturbations that are effective against the model but hardly affect human observers.
As previously described in Figure \ref{fig:fusion}, we generate a semantic-preserved adversarial example for the given image sample, and introduce an explanation consistency metric to measure the similarity between the model's visual explanations of the original sample and the semantic-preserved adversarial sample. 
We then assign a higher weight to the input image sample for further training if low explanation consistency is observed.

Our framework operates in an iterative manner during training. For each iteration, our framework consists of three stages, i.e., network training, explanation consistency evaluation, and sample reweighting. As illustrated in Figure~\ref{fig:framework}, our framework first trains the network in Stage 1. Then, we exploit the trained network to interpret the training samples and evaluating the explanation consistency between the training sample and its semantic-preserved adversarial example in Stage 2. Furthermore, we decrease the training weights of the samples exhibiting high explanation consistency and increase those with low explanation consistency for the next training iteration in Stage 3.

In the following, we illustrate how our framework works for improving the interpretability of the given network with the network parameters $\omega$. Here we introduce some notations to better describe our framework. We denote the recognition component of our framework as the ``performer'' $f(\cdot)$  and the interpreting component of our framework as the ``explainer'' $g(\cdot)$. Note that, for the network visualization methods, the performer and the explainer are decoupled, i.e., the target CNN to be interpreted is termed as the performer while the network visualization methods used to interpret the target CNN is termed as the explainer. 
For interpretable networks, the performer and explainer are highly integrated and need to be learned jointly.

Following previous works, we focus on the image classification task. Suppose we have $n$ samples $\{x_i\}_{i=1}^n$ that are taken from $C$ distinct categories. Here, $x_i$ denotes the $i$-th input image, and its associated label is $y_i \in \{1, ..., C\}$. Given the performer $f(\cdot)$ and the explainer $g(\cdot)$, we thus have the label prediction $f(x_i)$ and semantic heatmap $g(f(x_i))$ for each image $x_i$. Note that, semantic heatmap $g(f(x_i))$ shares the same image size with $x_i$ and is a probability map, which highlights the important semantic pixels of the image by the explainer. For network visualization methods, the coarse localization map $g(f(x_i))$ that highlights important regions is generated by flowing the gradients of the predicted label into the final convolutional layer, while $g(f(x_i))$ implies the top convolutional layers of a CNN in some interpretable network methods (e.g., \cite{icnn}). Most CAM-based visualization methods are built on the top layer of a CNN with normalization and ReLU activation, providing heatmaps with values ranging from 0 to 1. In cases where $g(f(x_i))$ does not naturally adhere to this range, we employ clipping to enforce it.

\subsection{Semantic-preserved Adversarial Attack}
In order to evaluate the interpreting robustness of the explainer towards the given sample $x_i$, we leverage the image adversarial attack techniques to generate the adversarial perturbation/noise of $x_i$, which aims to challenge the performer. We let $h(\cdot)$ denote the adversarial function and $h(x_i, y_i)$ denote the initial adversarial perturbation for $x_i$ and its label $y_i$. Then, we regard the semantic heatmap $g(f(x_i))$ as a mask and filter out those pixels with semantic meanings in it, i.e., $\mathbf{1}- g(f(x_i))$. Therefore, we have the semantic-preserved adversarial example $\hat{x}_i$ as:
\begin{equation}
\label{eq:adver}
    \hat{x}_i = x_i + h(x_i, y_i) \circ (\mathbf{1}- g(f(x_i))),
\end{equation}
where $\circ$ denotes the Hadamard product. Note that $\hat{x}_i$ will be only used in computing \textit{explanation consistency}. The key idea of the equation is to attack the regions with less semantic meaning. As depicted in Figure~\ref{fig:fusion}, the adversarial perturbation $h(x_i, y_i)$ takes a greater effect on those image regions with less semantic meaning corresponding to the semantic heatmap $g(f(x_i))$. So the more explainable a semantic heatmap is, the smaller the performance drop caused by the perturbation (i.e., $f(x_i)\approx f(\hat{x}_i)$) since the attack is barely focused on the semantic regions that are less correlated with its prediction.

\begin{algorithm}[!t]
\caption{Model Training with Our Proposed Framework}
\label{alg:alg_overview}
\begin{algorithmic}[1]
\REQUIRE Training dataset $\{{x}_{i}\}_{i=1}^{n}$ with the corresponding labels $\{{y}_{i}\}_{i=1}^{n}$
\ENSURE Model parameters $\omega$ 
\STATE Initialize network parameters $\omega$ and the training weights $\{{v}_{i}\}_{i=1}^{n}={1}$ ;
\STATE \textbf{for} Iteration = $1,...,T$ \textbf{do}
\STATE ~~ Perform network training to update $\omega$ via Eq.~(\ref{eq:obj}) ;
\STATE  ~~ Generate semantic-preserved adversarial examples \\
$\{\hat{x}_{i}\}_{i=1}^{n}$ via Eq.~(\ref{eq:adver}) ;
\STATE ~~~ Evaluate the explanation consistency to update $\{{v}_{i}\}_{i=1}^{n}$ via Eq.~(\ref{eq:v})  ;
\STATE \textbf{end for}
\RETURN $\omega$ ;
\end{algorithmic}
\end{algorithm}

\subsection{Interpretability Learning by Evaluating Explanation Consistency}
We further feed the semantic-preserved adversarial example $\hat{x}_i$ into the network to produce its semantic heatmap $g(f(\hat{x}_i))$ and label predictions $f(\hat{x}_i)$. Then, we consider the explanation consistency between the input image $x_i$ and its semantic-preserved adversarial example $\hat{x}_i$. Specifically, in an interpretable network, a semantic heatmap $g(f(\cdot))$ is supposed to be highly relevant with the prediction $f(\cdot)$. Thus, $x_i$ and $\hat{x}_i$ not only should share their predictions ($f(x_i)\approx f(\hat{x}_i)$), but also should maintain the close explanations ($g(f(x_i))\approx g(f(\hat{x}_i))$) toward their predictions.

Therefore, we define the explanation consistency metric as follows:
\begin{equation}
\label{eq:metric}
 E(x_i, \hat{x}_i)= \left \{
\begin{aligned}
 & 1-D(x_i, \hat{x}_i),~ & f(\hat{x}_i) = f(x_i), \\
 & 0, & f(\hat{x}_i) \not = f(x_i), \\
\end{aligned} 
\right.
\end{equation}
\begin{equation}
\label{eq:dis}
 D(x_i, \hat{x}_i)= \psi(\big||g(f(\hat{x}_i)) - g(f(x_i)) ||_2^2 \big), \\
\end{equation}
where $\psi(\cdot)$ is a function that maps positive numbers into the interval $[0,1]$ in a monotonically increasing manner. Here, we utilize the hyperbolic tangent function $\tanh(\cdot)$ for this purpose.

Note that, we directly assign the lowest consistency score, i.e., $E(x_i, \hat{x}_i)=0$, when the predictions $f(\hat{x}_i)$ and $f(x_i)$ are different. That is, the consistency score $E(x_i, \hat{x}_i)$ becomes minimal if the attack changes the prediction of $x_i$. This usually implies the semantic heatmap $g(f(x_i))$ is not satisfactory due to the limitation of network interpretability. On the other hand, if the prediction does no change, the consistency score resorts to the variation of their explanations and measures the difference between $g(f(\hat{x}_i))$ and $g(f(x_i))$.

Under this explanation consistency rule, we develop a new sample-reweighting strategy that simultaneously improves the model's performance and interpretability:
\begin{equation}
\label{eq:v}
 v_i = 1 - E(x_i, \hat{x}_i) = \left \{
\begin{aligned}
 & D(x_i, \hat{x}_i),~ & f(\hat{x}_i) = f(x_i), \\
 & 1, & f(\hat{x}_i) \not = f(x_i), \\
\end{aligned} 
\right.
\end{equation}
where $v_i\in(0,1]$ reflects the explanation inconsistency of a sample $x_i$. For $x_i$ with lower explanation consistency, a higher weight $v_i$ will be assigned to it, thereby making it receive more attention in model training.

By applying Eq. (\ref{eq:v}) to each sample $x_i$ and obtaining the training weight set $\textbf{v} = \{v_i\}_{i=1}^n$, we formulate our overall loss function as: 
\begin{equation}
\label{eq:obj}
Loss = \sum_{i=1}^n v_i L_i(x_i, y_i),
\end{equation}
where $L_i(x_i, y_i)$ denotes the standard cross entropy loss for each sample $x_i$, defined as:
\begin{equation}
\begin{split}
    L_i(x_i, y_i) = - \big( y_i \log f(x_i) + (1 - y_i)\log (1 - f(x_i) ) \big).
\end{split}
\end{equation}
Note that, Eq. (\ref{eq:obj}) naturally coincides with the hard negative mining strategy \cite{ohem}. However, in our framework, more difficult samples are related to lower explanation consistency. Our framework assigns higher weights to samples with lower explanation consistency in Eq. (\ref{eq:obj}), so as to endow a model with good performance and explanation concurrently. During the training, we directly update the network parameters $\omega$ by the standard back-propagation. 

The entire pipeline of training a model with our proposed framework is summarized into Algorithm \ref{alg:alg_overview}.

\begin{figure*}[!t]
\centering
\includegraphics[width = 0.98\textwidth]{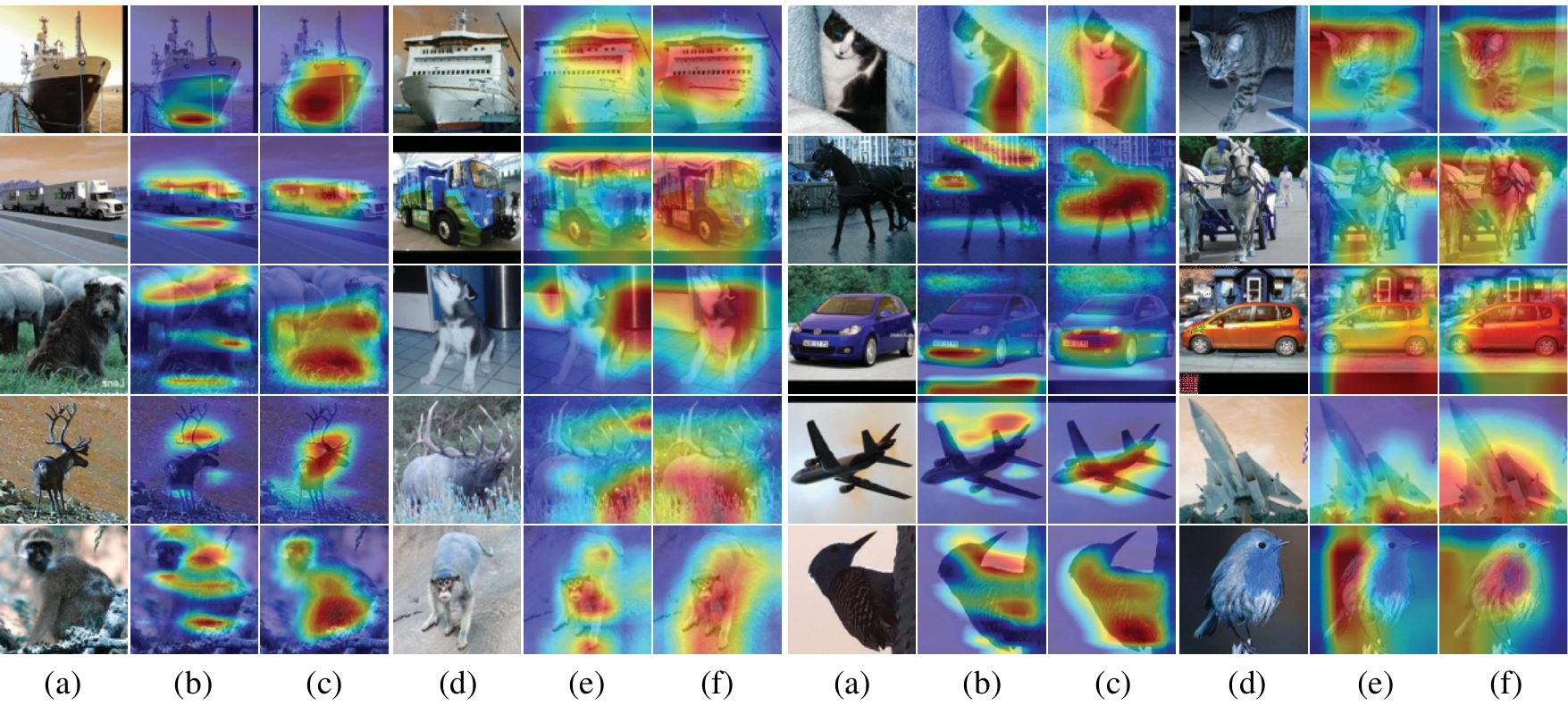}
\caption{Qualitative comparison of explanation maps on the STL-10 dataset. (a) and (d) denote the input images; (b) and (e) denote the explanation results generated by the Grad-CAM method on the regularly trained VGG-16 and ResNet-50 models, respectively; (c) and (f) denote the explanation results generated by Grad-CAM on the VGG-16 and ResNet-50 models that are trained via our framework. As shown, the visualization method gives more accurate semantic localization. The results imply that models trained with our framework have better interpretability than the baseline models.} 
\label{fig:vgg_resnet}
\end{figure*}

\begin{table*}[tbp]
\centering
\small
\caption{A quantitative analysis on PASCAL VOC object segmentation dataset. The best and second best results are highlighted with bold and underlined fonts, respectively.}
\label{tab:result2}
\begin{tabular}{c|c|c|c|c|c}
\hline 
\hline 
\multirow{2}*{Method}   & \multirow{2}*{Network}  & \multirow{2}*{Accuracy}  & \multicolumn{3}{c}{Interpretability (mIoU)} \\
& &                      & XGrad-CAM \cite{xgradcam}         & Grad-CAM \cite{gradcam}        & Ablation-CAM \cite{ablationcam} \\
\hline
\hline

Baseline      & \multirow{6}*{ResNet-50 \cite{resnet}}   & 45.72\%          & 0.0938 & 0.0938 & 0.1239    \\
HNM \cite{ohem}             &              & 44.89\%          & 0.1198 & 0.1198 & 0.1402    \\
FL \cite{lin2017focal}      &              & 45.10\%          & 0.1046 & 0.1042 & 0.1236    \\
EFL \cite{li2022equalized}  &              & 44.54\%          & 0.1067 & 0.1068 & 0.1174    \\
Ours-XGrad-CAM               & & \underline{46.41\%} & \textbf{0.1285} & \textbf{0.1285} & \underline{0.1456}  \\
Ours-Eigen-CAM               & & \textbf{46.48\%} & \underline{0.1254} & \underline{0.1254} & \textbf{0.1498}  \\
\hline
Baseline      & \multirow{6}*{MobileNetV2 \cite{sandler2018mobilenetv2}} & 53.31\%             & 0.1967 & 0.1967 & 0.2123    \\
HNM \cite{ohem}             &              & 53.69\%             & 0.1971 & 0.1971 & 0.2106    \\
FL \cite{lin2017focal}      &              & 53.38\% & \underline{0.1988} & \underline{0.1988} & 0.2112    \\
EFL \cite{li2022equalized}  &              & 53.45\%             & 0.1973 & 0.1973 & 0.2099    \\
Ours-XGrad-CAM               & & \textbf{55.80\%}     & \textbf{0.1990} & \textbf{0.1990} & \underline{0.2125}   \\
Ours-Eigen-CAM               & & \underline{55.32\%}  & 0.1974 & 0.1975 & \textbf{0.2128}  \\
\hline
\hline
\end{tabular}\vspace{12pt}
\end{table*}

\subsection{Implementation and Discussion}

\textit{Implementation of Adversarial Attack.}~~In all our experiments, we utilize the image adversarial attack technique FGSM \cite{ehae} to generate $h(x_i)$ due to its wide usage and easy implementation. In particular, $h(x_i)$ is formulated as:
\begin{equation}\label{eq:adv}
h(x_i, y_i)= \epsilon \cdot {sign}(\bigtriangledown{x_i} L(x_i, y_i)),
\end{equation}
where $\bigtriangledown{x_i}$ denotes the gradient of the given image $x_i$ and $\epsilon$ represents the perturbation level and is empirically set as 0.1. The function ${sign}(\cdot)$ is used to extract the sign of numeric values. Specifically, it returns $-1$ for negative numbers and 1 for non-negative numbers.

\textit{Extension to multi-label image classification.}~~Our framework primarily focuses on single-label image classification. However, it can be directly applied to the multi-label image classification task \cite{ChenXHWL19iccv,PuLWCTLL24TMM} in a simple manner: for a given training sample, consider only the highest-confidence predicted label to evaluate explanation consistency and reweight the sample accordingly. If we consider multiple predicted labels for a sample in explanation consistency evaluation, further investigation is needed to determine how to generate an appropriate explainable heatmap and ensure that the sample reweighting is effective for these predicted labels.

\textit{Extension to different architectures and unseen data.}~~This work focuses on improving the interpretability of convolutional neural network architectures. Our framework could potentially be extended to different network architectures (e.g., Transformer-based models \cite{DosovitskiyB0WZ21ViT,PuCWLL24}), but this would require exploring specific network visualization methods tailored to those architectures for explanation consistency evaluation. Additionally, since our framework only reweights samples during model training, integrating it with existing effective transfer learning methods can help mitigate the risk of overfitting on limited data scenarios and enhance the model's generalization capability on test data from new scenarios. Furthermore, visual-language models (VLMs) have recently demonstrated strong transfer learning abilities \cite{RadfordKHRGASAM21icml,Wu2023PersonSearch,Cheng0HLHP18mm,OhKLPJ2023NIPSW,WangSZJW0W21cvpr}, providing a better understanding of the open real world \cite{LiuMZWGY24pami,HuangCLW2018NIPSW,HeGDQS0X23aaai}. Our framework could also be extended to combine with VLMs to improve network interpretability for unseen categories. In such cases, it may be essential to consider adding additional constraints during the training of VLMs.

\textit{Connection with counterfactuals.}~~The concept of counterfactuals refers to the causal explanation about what would have been true by changing the circumstances \cite{pearl2018book}. Interestingly, we found our method has some intrinsic relation with counterfactuals. Concretely, let's treat the semantic-preserved adversarial attack as a sort of independent treatment from the feature map to the prediction.  If unactivated regions in a feature map can not explain the network's outcome, the manipulation to the regions would not change the prediction. To this, chasing for evaluation consistency may refer to the counterfactual fairness before and after the manipulation. It provides an explanation about why a simple re-weighting scheme may help learn explainable heat maps.

\section{Experiments}\label{sec:Experiments}
To validate the effectiveness of our proposed framework in improving the network's performance and interpretability, we conduct extensive experiments with different visualization methods \cite{xgradcam,gradcam,ablationcam} on both regular \cite{resnet,sandler2018mobilenetv2} and interpretable networks \cite{iccnn}. 

i) For regular networks, we tested on the {STL-10} \cite{pmlr-v15-coates11a} and {PASCAL VOC 2012 Object Segmentation (VOC)} benchmarks \cite{pascal-voc-2012}. STL-10 consists of 5000 training images for 10 classes and 800 test images per class. To quantify the improvement achieved by our framework, we introduce the VOC benchmark which provides pixel-wise object segmentations for 20 classes.

ii) In Section \ref{sec:Improving Intepretable Networks}, having our framework building on an interpretable network, icCNN \cite{iccnn}, we first evaluate our approach on the public {Caltech-UCSD Birds-200-2011 (CUB-200-2011) Benchmark} \cite{wah2011caltech}, which contains 200 bird categories. To further examine our framework's robustness to background shifts, we conduct experiments on ImageNet-9 \cite{imagenet9}. ImageNet-9 is a synthetic perturbed background dataset of nine categories, each including 5045 training samples and 450 test samples, with a variety of blending of foregrounds and backgrounds aiming at encouraging deep networks to disentangle foreground and background features. 

iii) Further, we evaluate our framework and the compared methods on the condition of insufficient training data. We also evaluate the model's debiasing capability via reweighting and demonstrate that the choice of using adversarial perturbed samples in the reweighting process can help improve the model robustness.

\begin{table*}[!t]
\setlength{\tabcolsep}{8pt}
\renewcommand{\arraystretch}{1.1}
\small
\centering
\caption{Human evaluation results on the quality of visual explanations from the models trained via our framework, obtained through Amazon Mechanical Turk (MTurk). The ``pos'', ``same'' and ``neg'' columns represent the percentages of examples that human raters perceived as an improvement, no substantial changes, or a setback, respectively.}
\label{tab:human}
\begin{tabular}{c|c|c|c|c|c|c|c}
\hline
\hline
\multirow{2}{*}{Network} & \multirow{2}{*}{Method} &  \multicolumn{3}{c|}{STL-10} & \multicolumn{3}{c}{VOC} \\
& & pos & same & neg & pos & same & neg \\
\hline
\hline
\multirow{2}{*}{VGG-16 \cite{vgg}} & Grad-CAM \cite{gradcam} & 37\% & 53\% & 10\% & 39\% & 50\% & 11\%\\
 & Inte-Grad \cite{integrad} & 34\% & 52\% & 14\%& 38\% & 46\% & 16\%\\
\hline
\multirow{2}{*}{ResNet-50 \cite{resnet}} & Grad-CAM \cite{gradcam} & 43\% & 49\% & 8\% & 59\% & 30\% & 11\% \\
 & Inte-Grad \cite{integrad} & 45\% & 45\% & 10\% & 62\% & 28\% & 10\%\\
\hline
\hline
\end{tabular}
\end{table*}

\begin{figure*}[!t]
\centering
\includegraphics[width = 0.96\textwidth]{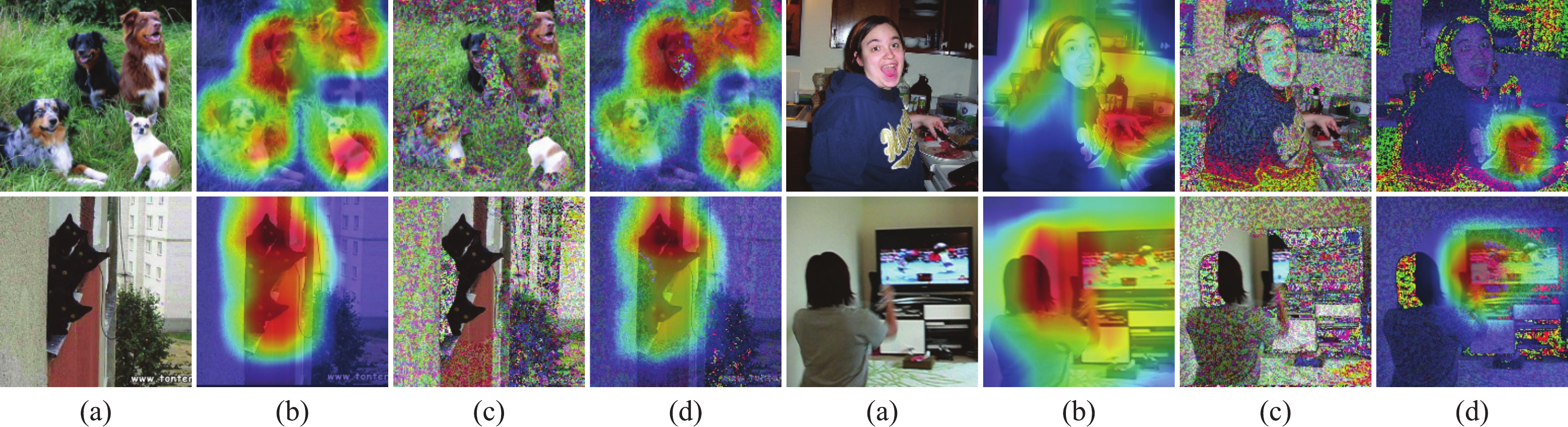}
\caption{Some examples with high (left) or low (right) explanation consistency from the STL-10 benchmark, obtained by the ResNet-50 model trained with our framework. (a) the input image; (b) the initial explanation result; (c) the semantic-preserved adversarial image; (d) the new explanation result.} 
\label{fig:hard}
\end{figure*}

\subsection{Improving Networks via Different Visualization Methods}\label{sec:VisualizationMethods}
We build our framework using various post hoc visualization methods, including {Grad-CAM} \cite{gradcam}, {Inte-Grad} \cite{integrad}, Axiom-based Grad-CAM (XGrad-CAM) \cite{xgradcam} and Eigen-CAM \cite{muhammad2020eigen}, on the benchmarks.
We use the class labels as supervision in the training process and use the pixel-wise annotations only to evaluate interpretability.
We follow previous works to use the mean Intersection-over-Union (mIOU) score as the evaluation metric for interpretability.

\subsubsection{Experimental Details}
We first train both the VGG-16 and ResNet-50 models with the STL-10 dataset, respectively. The batch size is 100, while the initial learning rate is 0.0001, with a step learning rate decayed by 0.1 every 30 epochs. The Adam optimizer \cite{kingma2014adam} is employed. The class activation map's threshold is empirically set as 0.5 in our experiments. As outlined above, we measure the explanation consistency of the training samples before and after the image adversarial attack FGSM \cite{ehae}.
On the VOC benchmark, we extract multi-class labels based on objects that appear in the segmentation and use these labels to train a classification model. The experiment setup is nearly the same as above, except that: (i) we have a lower batch size of 8; (ii) we further compare the pixel-wise segmentation ground truth with heatmaps produced by the trained models. Specifically, we compute the mIoU scores between the ground truth regions and thresholded activation maps generated from three different network visualization methods: XGrad-CAM \cite{xgradcam}, Grad-CAM \cite{gradcam}, and Ablation-CAM \cite{ablationcam}. The reported results are are the average of three trials. 

\begin{table}[!tbp]
\small
\renewcommand{\arraystretch}{1.0}
\centering
\caption{Human evaluation results on the quality of visual explanations from the models trained via HNM and our framework, obtained through Amazon Mechanical Turk (MTurk). The ``pos'', ``same'' and ``neg'' columns represent the percentages of examples that human raters perceived as an improvement, no substantial changes, or a setback, respectively.}
\label{tab:human_hnm}
\begin{tabular}{c|c|c|c|c}
\hline
\hline
\multirow{2}{*}{Network} & \multirow{2}{*}{Method} &   \multicolumn{3}{c}{VOC} \\
& & pos & same & neg \\
\hline
\hline
\multirow{2}{*}{ResNet-HNM \cite{ohem}} & Grad-CAM \cite{gradcam} & 38\% & 21\% & 41\%\\
 & Inte-Grad \cite{integrad} & 42\% & 6\% & 52\%\\
\hline
\multirow{2}{*}{ResNet-Ours} & Grad-CAM \cite{gradcam} & 59\% & 30\% & 11\% \\
 & Inte-Grad \cite{integrad} & 62\% & 28\% & 10\%\\
\hline
\hline
\end{tabular}
\end{table}

\begin{figure*}[!t]
\centering
\includegraphics[width = 0.88\linewidth]{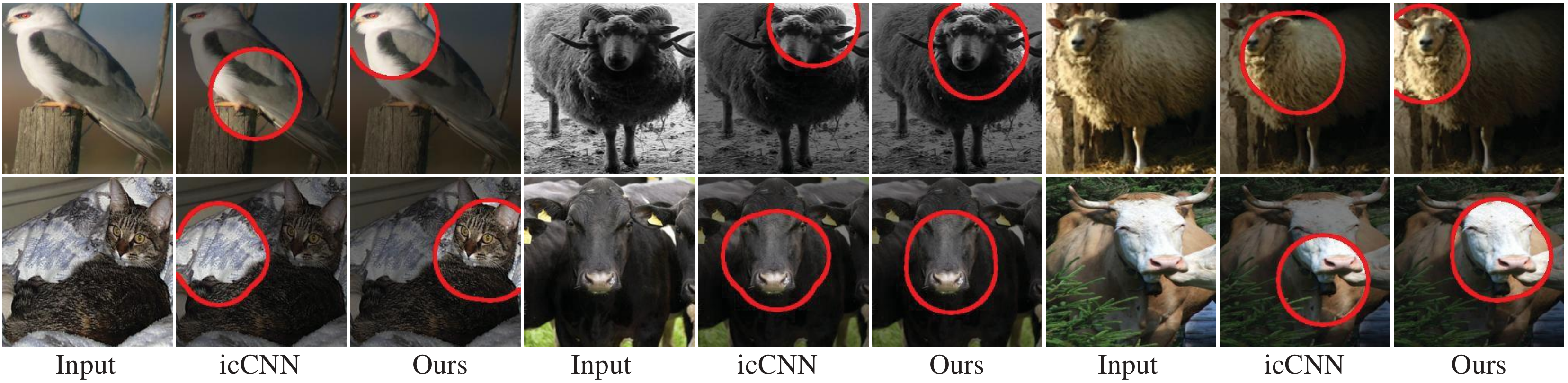}
\caption{Qualitative comparison of visualization of filters in top convolutional layers of an interpretable CNN (icCNN). Visualizations are generated following the setup in \cite{icnn} to ensure a fair comparison.}
\label{fig:visualcmp}
\end{figure*}

\begin{figure}[!t]
\centering
\includegraphics[width = 0.99\linewidth]{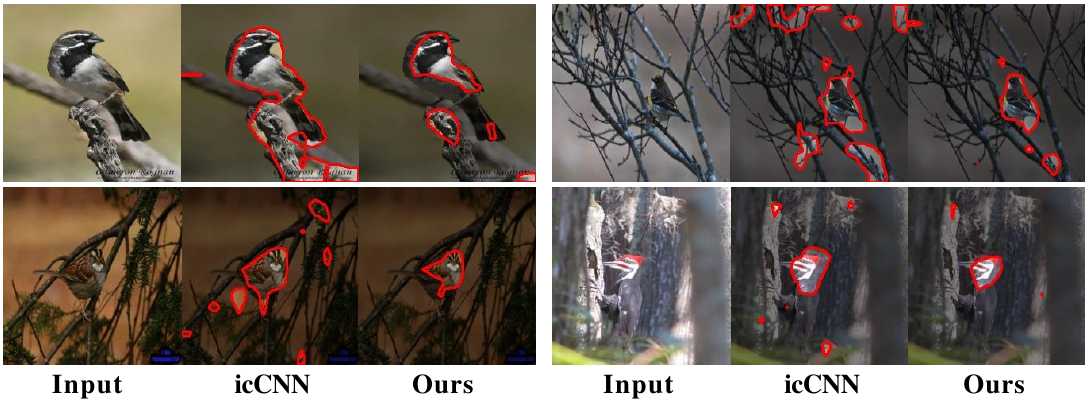}
\caption{\small{Qualitative comparison of visualization of interpretable filters in the interpretable compositional CNN (icCNN) \cite{iccnn}. The network trained with our framework shows more precise object localization power and better robustness in locating foreground objects against complex backgrounds.}}
\label{fig:visualcmpContour}
\end{figure}

\subsubsection{Results and Analyses}

Since the STL-10 dataset does not include segmentation ground truth for mIoU calculation, we provide qualitative results and analysis instead.  Figure~\ref{fig:vgg_resnet} illustrates the qualitative comparison of Grad-CAM explanations between the baseline model and the reweighted version by our framework. The baseline models are the regularly trained VGG-16 and ResNet-50 networks. It can be observed that the models trained with our framework produce superior explainable heatmaps.

Table \ref{tab:result2} shows the accuracy and mIoU results on the VOC dataset. The baseline models are the regularly trained ResNet-50 \cite{resnet} and MobileNetV2 \cite{sandler2018mobilenetv2} networks. We utilize two different network visualization methods XGrad-CAM \cite{xgradcam} and Eigen-CAM \cite{muhammad2020eigen} in our framework as the explainer, and the respective versions of our method are denoted as ``Ours-XGrad-CAM'' and ``Ours-Eigen-CAM'' in Table \ref{tab:result2}. Besides the baseline, we further compare our framework with a range of existing reweighting methods, including hard negative mining (HNM) \cite{ohem}, focal loss (FL) \cite{lin2017focal}, and equalized focal loss (EFL) \cite{li2022equalized}. In our experiments for the counterparts, we maintain the same experimental settings as ours to ensure a fair comparison. As displayed in the upper part of Table~\ref{tab:result2}, when associated with ResNet-50 \cite{resnet}, our framework consistently surpasses existing methods by large margins in terms of accuracy and mIoU. Significantly, our framework leads to a notable increase of about 1\% in accuracy and about 7\% in relative interpretability compared to HNM, which stands as the best among our counterparts. For MobileNetV2 \cite{sandler2018mobilenetv2}, as shown in the lower part of Table~\ref{tab:result2}, our framework outperforms others in accuracy by a margin of 1.63\% and maintains its competitive edge in mIoU.

As interpretability is subjective, mIoU scores may not be a perfect evaluation metric. Therefore, we also ask human raters on Amazon Mechanical Turk (MTurk) to help evaluate the explanations. The raters are given two visual explanations of the same image: one is given by the regularly trained model, the other is given by the model trained via an enhanced method (e.g., HNM \cite{ohem} or our framework). Two explanations have an equal possibility to be shown as the first or the second to avoid potential bias. We also explain how to understand a heatmap and give examples of better versus worse explanations. 
Then the raters are asked to choose one explanation that seems to be more informative to them compared to the other or decide the two explanations do not have a substantial distinction.  For each dataset, we randomly select 100 testing images. There is no limit of the maximum number of tasks a worker can perform. For each pair of explanations, we collect the evaluations from at least 5 different workers. Finally, we decide the human opinion for each pair of explanations based on the majority votes. If there is a tie between the two explanations, we consider two explanations do not have substantial differences.

By applying our proposed framework to the VGG-16 \cite{vgg} and ResNet-50 \cite{resnet} models, Table~\ref{tab:human} reports the percentages of examples that human raters perceive as an improvement (denoted as ``pos''), no substantial changes (denoted as ``same''), or a setback (denoted as ``neg''), respectively, on the quality of visual explanations.
As shown in Table~\ref{tab:human}, our proposed framework significantly improves the quality of visual explanations in all tested settings. Similarly, the results with Inte-Grad consistently show comparable improvements. We also ask human raters to compare our framework with HNM and the results are in Table~\ref{tab:human_hnm}. While HNM can help improve the model's performance, the quality of the explanations provided by models trained with HNM is inconsistent. This observation confirms the mIoU results in Table~\ref{tab:result2} that our framework stabilizes the model and enables it to ferret out interpretable features. Figure~\ref{fig:hard} shows some examples with low explanation consistency versus high explanation consistency from the STL-10 benchmark. As observed in the right half of the figure, the visual heatmaps of samples with low explanation consistency differ significantly from those of their semantic-preserved adversarial counterparts.

\begin{table}[!t]
\centering
\small
\caption{Evaluation results of an interpretable network on CUB-200-2011 (CUB) and ImageNet-9 (IN-9). }
\label{tab:interp}
\begin{tabular}{c|c|c|c|c}
\hline
\hline
\multirow{2}{*}{Method}  & \multirow{2}{*}{Backbone}  & \multicolumn{2}{c}{CUB} & IN-9  \\
                         &                            & Accuracy & mIoU          & Accuracy    \\
\hline
\hline
Baseline    & \multirow{4}*{icCNN \cite{iccnn}}     & 66.2\% & 21.87\% & 70.7\%  \\
HNM \cite{ohem}         &                           & 71.8\% & 22.34\% & 78.8\%  \\ 
AL \cite{al}            &                           & -      & - & 80.1\%  \\
Ours                    &                           & \textbf{73.4\%} & \textbf{22.52\%} & \textbf{82.1\%} \\
\hline
\hline
\end{tabular}
\end{table}

\subsection{Improving Intepretable Networks} \label{sec:Improving Intepretable Networks}

In addition to the general network, we also evaluate our framework on an interpretable network, icCNN \cite{iccnn}, to verify the framework's generalizability on different kinds of networks. We compare our framework with related methods \cite{al, ohem} on the CUB-200-2011 benchmark. Besides, we challenge the framework on ImageNet-9 \cite{imagenet9} under an out-of-distribution background scenario.

\subsubsection{Experimental Setting}
We continue to adopt mIoU as the evaluation metric of network interpretability, with classification accuracy also reported. We follow the training procedure of icCNN \cite{iccnn}, except for adjusting the weights of training samples based on explanation consistency evaluation. The training configurations remain unchanged on both datasets. The icCNN's filter partitions for different parts and large regions are optimized every two epochs. The loss coefficients $\lambda$ for learning part-specific filters and $\beta$ for learning category-specific filters \cite{iccnn} are both set to 0.1. The reweighting method is applied after every four epochs. Similar to previous experiments, we use the same size of training data for our framework and the compared methods to ensure a fair comparison.

\subsubsection{Results and Analyses}
Table~\ref{tab:interp} presents the experimental comparison among the baseline, HNM \cite{ohem}, AL \cite{al}, and our framework. As shown, on CUB-200-2011, while HNM boosts the baseline by 0.47\% in mIoU, our framework achieves the highest interpretability score (mIoU) over other counterparts. Also, our framework notably boosts the accuracy of the interpretable baseline by 7.2\%, surpassing HNM by 1.6\%. This observation implies that our framework can significantly improve interpretable networks' discriminative power and network interpretability. On ImageNet-9, our framework also notably enhances the accuracy of the interpretable baseline and consistently outperforms its counterparts by a clear margin. 
More importantly, our framework is more robust to background shifts and able to provide invariant predictions.

Figures~\ref{fig:visualcmp} and \ref{fig:visualcmpContour} exhibit the visual comparisons between our framework and the original interpretable network icCNN. In Figure~\ref{fig:visualcmp}, the explanation regions of the images are displayed at coarse granularity through red circles, while Figure~\ref{fig:visualcmpContour} outlines the explanation regions at fine granularity through contours. As shown, not only can our framework provide more precise red circles responding to the category semantic information, but it also enhances robustness against complex backgrounds.

\subsection{Ablation Study}

We conduct several ablation experiments to give more in-depth analysis of our framework. Specifically, we adopt the interpretable network icCNN \cite{iccnn} as the backbone network and evaluate on the CUB-200-2011 benchmark.

We first conduct an ablation experiment on the choice of the  mapping function $\psi(\cdot)$ in Eq. (\ref{eq:dis}). The function $\psi(\cdot)$ should map positive values to the interval $[0,1]$ in a monotonically increasing manner. Besides the hyperbolic tangent function $\tanh(\cdot)$ adopted in this paper, we also test another two functions that satisfy the requirements, i.e.,
$\text{softsign}(x)=x/(1+|x|)$ and $\text{arctan1}(x)=2\cdot\text{arctan}(x)/\pi$. 
The evaluation results are reported in Table \ref{tab:mapping}. It can be observed that the performances of using the three different mapping functions are comparable, indicating that the choice of mapping function does not have a significant impact. However, using $\text{tanh}(\cdot)$ yields slightly better performance than the other two in terms of both classification accuracy and interpretability, making it a reasonable choice in the experiments.

\begin{table}[!t]
\centering
\small
\caption{Evaluation results of an interpretable network with respect to different mapping functions $\psi(\cdot)$ on CUB-200-2011. }
\label{tab:mapping}
\begin{tabular}{c|c|c|c}
\hline
\hline
{Function}  & {Backbone}  & Accuracy & mIoU  \\
\hline
\hline
softsign    & \multirow{3}*{icCNN \cite{iccnn}}     & 73.1\% & 22.29\%\\
arctan1         &                           & 73.1\% & 22.05\% \\ 
tanh                    &                           & \textbf{73.4\%} & \textbf{22.52\%}\\
\hline
\hline
\end{tabular}
\end{table}

\begin{figure}[!t]
\centering
\includegraphics[width = 0.85\linewidth]{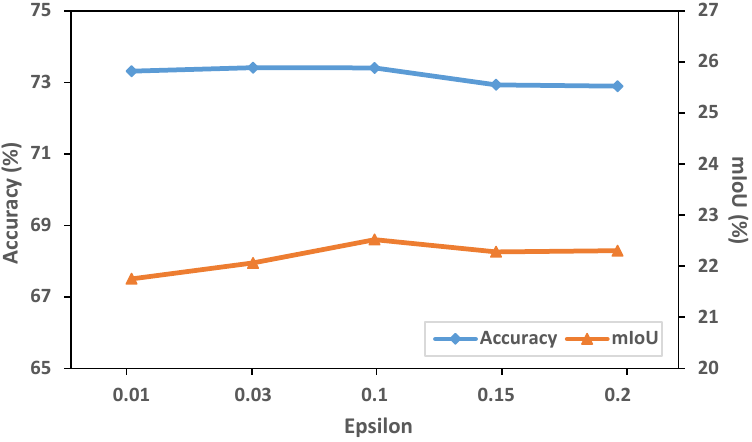}
\caption{Comparison of classification accuracy and interpretability (mIoU) with respect to different perturbation levels ($\epsilon$) on CUB-200-2011.}
\label{fig:epsilonCUB}
\end{figure}

We further conduct an experiment to evaluate the effect of the perturbation level (i.e., $\epsilon$ in Eq. (\ref{eq:adv})) on model performance. 
It just indirectly impacts model performance by affecting the weights of training samples since adversarial samples are not used in model training.  
The results are depicted in Figure \ref{fig:epsilonCUB}. It can be observed that, as $\epsilon$ gradually increases from 0.01, the accuracy remains nearly unchanged until it slightly decreases when $\epsilon$ exceeds 0.1. On the other hand, the mIoU score shows a slight increase followed by a slight decrease, with minor fluctuations. Overall, when $\epsilon$ is relatively small (less than 0.2), its variation just has a minor impact on the model performance. The results also suggest that the model achieves nearly optimal performance for both accuracy and mIoU when $\epsilon$ is around  0.1. Hence, in our experiments, we set the perturbation level $\epsilon=0.1$ for the FGSM adversarial attack \cite{ehae}.

Compared to the regular training scheme, our framework differs only in that we reweight the training samples by generating semantic-preserved adversarial samples and evaluating the explanation consistency.  Consequently, this reweighting process (including Stages 2 and 3 in our pipeline) incurs additional computational costs compared to the regular training scheme. Multiple experimental assessments indicate that the computational time of the reweighting process in our framework is five times more than that of the compared reweighting method HNM \cite{ohem}. The computational cost largely stems from generating semantic-preserved adversarial samples using FGSM \cite{ehae}. More sophisticated techniques are needed to accelerate this process.

\begin{figure*}[!t]
\centering
\includegraphics[width = 0.88\textwidth]{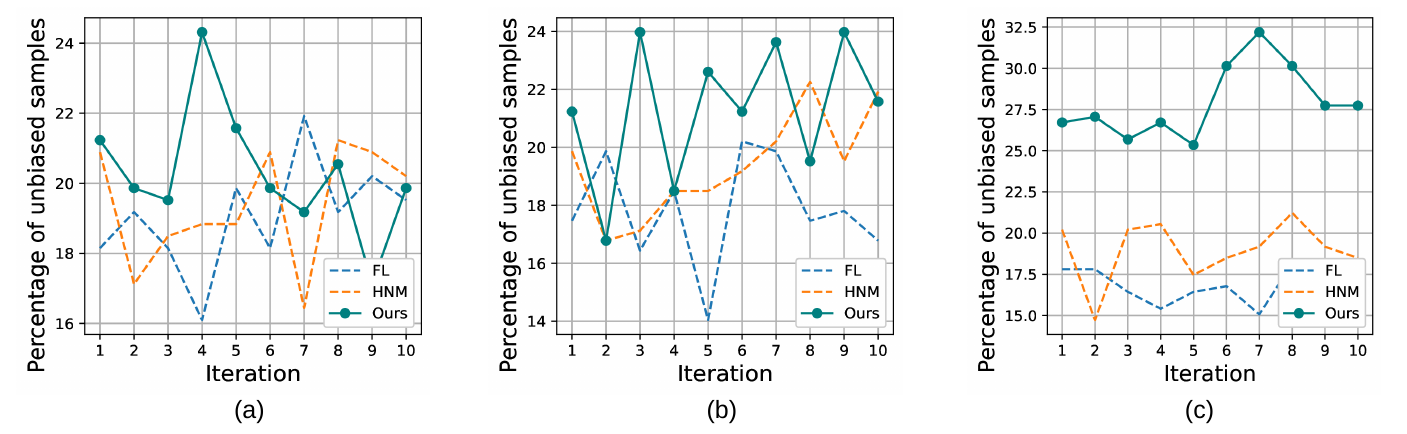}
\caption{\small{Comparison of different reweighting methods' debiasing power when different manual biases are involved in training data: (a) the brightness of 80\% samples is increased; (b) the brightness of 80\% samples is decreased; (c) the magnitude of 80\% samples is scaled up. Percentages of unbiased samples in the top 20\% of reweighting list are reported as a metric standing for debiasing power. }}
\label{fig:bias}
\end{figure*}

\begin{figure*}[!t]
\centering
\includegraphics[width = 0.95\textwidth]{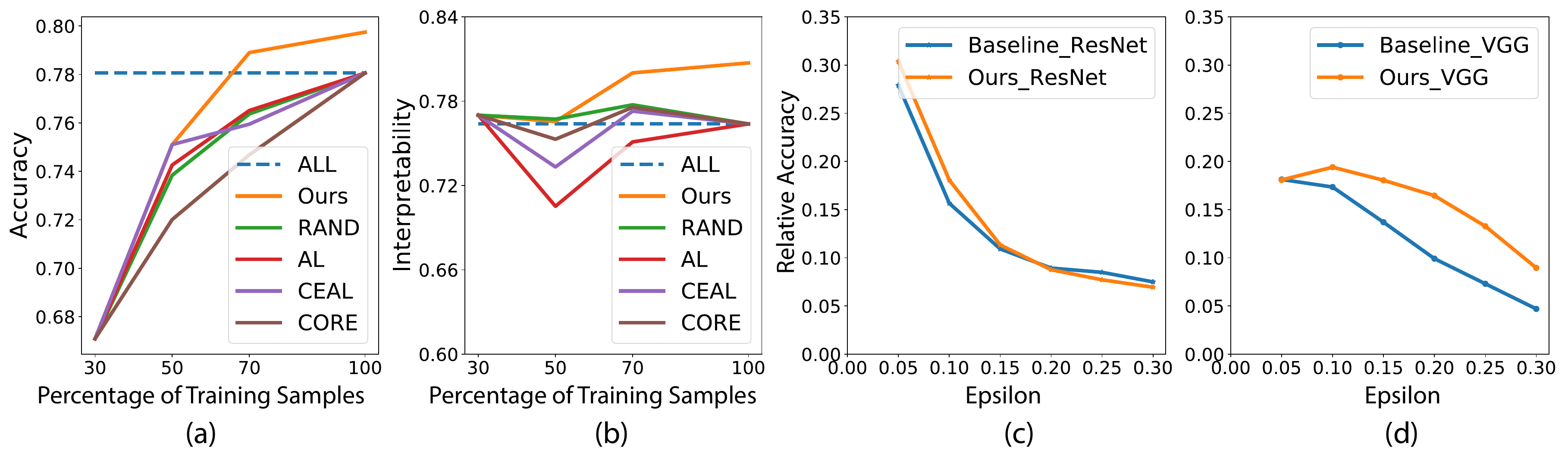}
\caption{Comparison of model learning with insufficient training samples and model robustness against perturbed samples. (a) and (b) show the quantitative comparison of classification accuracy and interpretability on the VOC Animal-Part val set, respectively. (c) and (d) demonstrate the relative accuracy of different models when facing model-specific FGSM adversarial attack at different perturbation levels ($\epsilon$) on ResNet and VGG, respectively. }
\label{fig:adv_resist}
\end{figure*}

\subsection{Evaluation of Debiasing Capability via Reweighting}
Reweighting or resampling is an effective measure to mitigate inherent biases in data \cite{amini2019uncovering}. To evaluate to what extent different reweighting approaches can debias data, we implant artificial biases in most of the training samples (80\%, in our experiments) and assess how many of the remaining samples (here referred to as unbiased samples for simplicity) are selected to counterweight the biased ones. Concretely, we train a MobileNetV2 for the PASCAL VOC benchmark in three artificial data bias settings: (i) the brightness of 80\% samples is increased by half their standard deviation; (ii) the brightness of 80\% samples is decreased by half their standard deviation; (iii) the magnitude of 80\% samples is scaled up by a factor of 1.5. We report the percentages of unbiased samples in the top 20\% of reweighting list ranked by FL \cite{lin2017focal}, HNM \cite{ohem}, and our framework for each iteration of reweighting until the training converges. 

Figure~\ref{fig:bias} shows the experimental results of the evaluation of debiasing capability. In Figure~\ref{fig:bias}(a) and Figure~\ref{fig:bias}(b), our framework captures about 24\% unbiased samples for training in multiple iterations in both setting (i) and setting (ii). At the same time, FL and HNM can only seize up to 22\% unbiased samples in a few iterations. Remarkably, Figure~\ref{fig:bias}(c) displays a significant margin between ours and other approaches, which even deliver, on average, less unbiased samples than random sampling (20\%). However, a larger number of unbiased samples are assigned higher weights by our framework in an attempt to eliminate the bias, which demonstrates the superiority of our framework in terms of data debiasing.

\subsection{Improving Model Learning from Insufficient Training Samples}\label{sec:InsufficientSamples}
To explore how the insufficiency of training samples affects network interpretability, we introduce a semi-supervised learning setting. Specifically, we first initialize the interpretable network by providing only 30\% annotations and allow the network to selectively add the remaining unused samples into training in an attempt to boost the recognition accuracy and interpretability. 

\subsubsection{Compared Approaches}
We compare our framework with two baseline methods, i.e., (i) ``RAND'', in which the remaining unused samples are randomly selected and added into training; (ii) ``ALL'', in which all the training samples are directly used for training. To demonstrate the superiority of our framework, we also include several state-of-the-art active learning and self-paced learning methods for comparison, i.e., (iii) ``AL'' \cite{al}, in which those samples with high prediction uncertainty are gradually added into training; (iv) ``CEAL'' \cite{ceal}, in which both high uncertainty samples and high certainty samples with pseudo-labeling are added into training; (v) ``CORE'' \cite{sener2017active}, in which the samples satisfy with the core-set selection are added for training.

\subsubsection{Results and Analyses}
For a fair comparison, we run our framework and all the compared methods three times and record the average accuracy and part interpretability in Figure~\ref{fig:adv_resist}. As shown in Figure~\ref{fig:adv_resist}(a), when given a gradually increasing number of training samples, our framework consistently outperforms all the compared methods by clear margins. It is worth mentioning that our framework even significantly performs much better than the ``ALL'' method, which directly uses 100\% training samples for training. Similarly, Figure~\ref{fig:adv_resist}(b) demonstrates that our framework also achieves the best interpretability as the number of training samples increases. These results justify the superiority of our framework over all the compared methods in terms of discriminative power and network interpretability.

In the experiments, we observe that the baseline ``RAND'' performs well in terms of interpretability compared to some curriculum learning-like methods (e.g., AL, CORE, and CEAL), as exhibited in Figure \ref{fig:adv_resist}(b), particularly when the percentage of training samples is below 70\%. Different from a typical curriculum learning setting where the model acquires most samples gradually during the entire training process, in this scenario, we train models only using a proportion of the training data. As discussed previously, training a model with insufficient samples can encourage the network to learn unreliable feature representations. This may explain why these methods do not perform well. It validates the advantage of our framework to address the weakness of such methods.

\subsection{Improving Model Robustness against Perturbed Samples}
We further conduct a simple experiment to justify that, for explanation consistency evaluation, using adversarial masked examples instead of zero/average-masked ones potentially makes trained models more resistant to adversarial attacks. For each model, we test with images under different degrees of adversarial perturbations using FGSM \cite{ehae}. Perturbations are created for each model independently based on its gradients on these images. For a fair comparison, we only use the test images that are correctly classified by both the baseline model and models trained using our proposed framework.

Figure~\ref{fig:adv_resist}(c) and Figure~\ref{fig:adv_resist}(d) illustrate how the accuracy changes on this sample subset (denoted as relative accuracy in the plots) for each model over different perturbation levels (i.e., $\epsilon$ in Eq. (\ref{eq:adv})). The results show that in general, the models trained with our framework perform better when facing model-specific FGSM adversarial attacks.

\section{Conclusion}\label{sec:Conclusion}
In this paper, we have proposed a concise framework for improving network interpretability by introducing the explanation consistency evaluation strategy. 
It reweights the training samples adaptively in model learning based on the explanation consistency metric, which measures the similarity between the model's visual explanations of the original samples and those of semantic-preserved adversarial samples.
Our framework highlights interesting connections between interpretability, adversarial examples, and curriculum learning. 
Extensive experimental evaluations on various benchmarks demonstrate the effectiveness of our proposed framework on both regular and interpretable networks. 

\section*{Acknowledgments}

The authors thank Yu Yang from UCLA for her valuable assistance in this work.

\bibliographystyle{IEEEtran}
\bibliography{reference}

\end{document}